\documentclass{article}

\usepackage{PRIMEarxiv}

\usepackage[utf8]{inputenc} 
\usepackage[T1]{fontenc}    
\usepackage{hyperref}       
\usepackage{url}            
\usepackage{booktabs}       
\usepackage{nicefrac}       
\usepackage{microtype}      
\usepackage{lipsum}
\usepackage{fancyhdr}       
\usepackage{graphicx}       
\graphicspath{{media/}}     

\usepackage{cite}
\usepackage{amsmath,amssymb,amsfonts}
\usepackage{algorithmic}
\usepackage{textcomp}
\usepackage{xcolor}
\def\BibTeX{{\rm B\kern-.05em{\sc i\kern-.025em b}\kern-.08em
    T\kern-.1667em\lower.7ex\hbox{E}\kern-.125emX}}
\usepackage{wrapfig}
\usepackage{color}
\usepackage{xspace}
\usepackage{multirow}
\usepackage{mathtools}
\usepackage[ruled]{algorithm2e}
\usepackage{textcomp}
\usepackage{wrapfig}
\usepackage{makecell}

\usepackage[table]{xcolor} 
\usepackage{array}        

\newcolumntype{P}[1]{>{\columncolor[gray]{0.96}}p{#1}}

\newcommand*{\figref}{Figure~\ref}
\newcommand*{\MRPs}{{MRPs}\@\xspace}

\newcommand*{\imu}{{IMU}\@\xspace}

\newcommand*{\rot }{ R }
\newcommand*{\quat}{ q }
\newcommand*{\mrp }{ \chi }

\newcommand*{\iimu}{I}
\newcommand*{\ilocal}{N}

\title{MinJointTracker: Real-time inertial kinematic chain tracking with joint position estimation and minimal state size
\thanks{These authors contributed equally to this work. This project has received funding from the European Union's Horizon Europe research and innovation programme under grant agreement No 101092889 (SHARESPACE), as well as the BMBF projects OrthoSuPer (project IDs 13GW0564A-F and 13GW0564E).}
}

\author{
Michael Lorenz$^{*}$ \\
Augmented Vision Group, German Research Center for Artificial Intelligence (DFKI) \\
Department of Computer Science, Rheinland-Pfälzische Technische Universität Kaiserslautern-Landau (RPTU) \\
Kaiserslautern, Germany \\
\texttt{michael.lorenz@dfki.de}
\And
Bertram Taetz$^{*}$ \\
IT \& Engineering, IU Internationale Hochschule \\
Erfurt, Germany \\
\texttt{bertram.taetz@iu.org}
\And
Gabriele Bleser-Taetz \\
IT \& Engineering, IU Internationale Hochschule \\
Erfurt, Germany \\
\texttt{gabriele.bleser-taetz@iu.org}
\And
Didier Stricker \\
Augmented Vision Group, German Research Center for Artificial Intelligence (DFKI) \\
Department of Computer Science, Rheinland-Pfälzische Technische Universität Kaiserslautern-Landau (RPTU) \\
Kaiserslautern, Germany \\
\texttt{didier.stricker@dfki.de}
}

\begin{document}
\maketitle

\begin{abstract}
Inertial motion capture is a promising approach for capturing motion outside the laboratory.
However, as one major drawback, most of the current methods require different quantities to be calibrated or computed offline as part of the setup process, such as segment lengths, relative orientations between inertial measurement units (IMUs) and segment coordinate frames (IMU-to-segment calibrations) or the joint positions in the IMU frames. 
This renders the setup process inconvenient.
This work contributes to real-time capable calibration-free inertial tracking of a kinematic chain, i.e. simultaneous recursive Bayesian estimation of global IMU angular kinematics and joint positions in the IMU frames, with a minimal state size. 
Experimental results on simulated IMU data from a three-link kinematic chain (manipulator study) as well as re-simulated IMU data from healthy humans walking (lower body study) show that the calibration-free and lightweight algorithm provides not only drift-free relative but also drift-free absolute orientation estimates with a global heading reference for only one IMU as well as robust and fast convergence of joint position estimates in the different movement scenarios.
\end{abstract}


\section{Introduction}
\begin{figure}[!b]
    \centering
    \includegraphics[width=0.70\columnwidth]{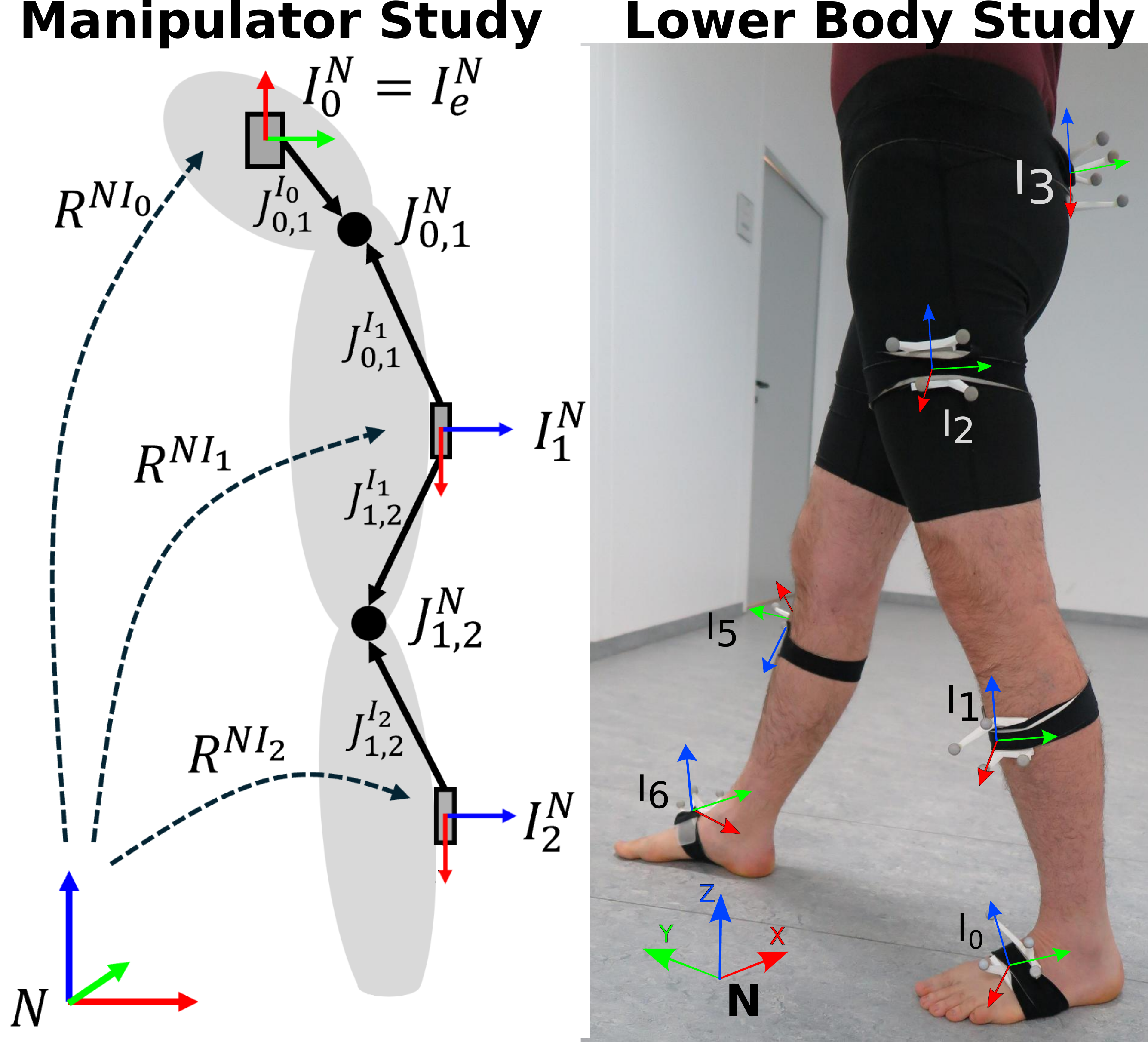}
    \caption{Left: Kinematic chain model with IMUs (grey rectangles), joint centers (black circles), vectors from IMU to joint center positions (solid black arrows) and IMU orientations (dashed black arrows). 
    The rigid segments, on which the IMUs are mounted, are indicated as light grey ellipses. 
    Right: Lower body real setup. The notation can be transferred from the illustration on the left side, and $I_3=I_e$.
        \label{fig:kinChain}
    }
\end{figure}

Inertial human motion tracking uses multiple inertial measurement units (IMUs) attached to body segments by means of straps or clothing integration \cite{Cereatti2024}. 
Motion in terms of segment or joint kinematics is typically deduced by using a suitable state estimation approach in combination with a personalized kinematic model \cite{García-de-Villa2023,Cereatti2024}. 
However, kinematics estimation comes with several challenges, such as integration drift, in particular when omitting the error-prone magnetometer information, and calibration.
The quantities to be calibrated are typically the segment lengths in order to obtain a personalized kinematic model, and the sensor poses with respect to the segment frames, the so-called IMU-to-segment calibrations. 
Their calibration is currently considered the main challenge in biomechanical motion analysis, and typical methods are based on manual and error-prone procedures~\cite{vitali2020}. 
This can reduce validity, reliability and usability of such a system.

In this work, we aim to achieve real-time capable calibration-free inertial tracking of the pose of a kinematic chain.
More specifically, we propose a recursive state estimation algorithm with minimal state size for simultaneous estimation of (1) IMU angular kinematics and (2) joint positions w.r.t. to the associated IMU coordinate frames in a kinematic chain.
Using the estimated IMU orientations, the estimated joint positions can then be transformed into a common reference frame and can be connected to illustrate the body pose.

Related work in this area comprises previous IMU-based approaches to 
\begin{enumerate}
    \item offline joint position estimation \cite{Seel2012,Skoglund2015,olsson2017},
    \item joint kinematics estimation assuming pre-estimated joint positions \cite{weygersDriftFreeInertialSensorBased2020}  and
    \item simultaneous estimation of IMU kinematics and joint positions \cite{McGrath2021,Taetz2024}.
\end{enumerate}
Concerning the latter, \cite{McGrath2021} is also offline and makes segment length assumptions, while \cite{Taetz2024} is most related to this work, but it uses a higher-dimensional state vector including position, velocity, acceleration, orientation and angular velocity for each IMU. 
In this work, we leverage the idea of drift-free joint orientation tracking with a minimal state as proposed in \cite{weygersDriftFreeInertialSensorBased2020} and apply it to \cite{Taetz2024}, which estimates joint positions online with a high-dimensional state.
Furthermore, we exploit \cite{Lorenz_Taetz_Bleser_Stricker_2022}, which allows us to use only one additional source of absolute orientation to align a kinematic chain with a global reference frame.
The main contributions of our proposed algorithm are the following:
\begin{itemize}
    \item It is calibration-free and capable of estimating drift-free relative orientations of IMUs attached to a kinematic chain.
    \item It needs only one source of global orientation or heading information for global alignment, which could be based on one IMU in the kinematic chain with magnetometers or any other sensor capable of providing drift-free absolute orientations with respect to one IMU. 
    \item It is able to compute joint positions online with minimal computational resources due to a minimal state representation.
\end{itemize}
The paper is structured as follows:
In Section~\ref{sec:methods} we introduce the complete method and experimental setup including a manipulator and a lower body study, as illustrated in \figref{fig:kinChain}.
The experimental results are presented in Section~\ref{sec:results} and are discussed in Section \ref{sec:discussion}.
Section \ref{sec:conclusions} presents conclusions and future work.


\section{Methodology}
\label{sec:methods}

\subsection{Notation and kinematic chain model}
\label{sec:biomechModel}
The kinematic chain model is exemplified with three IMUs and two joints in Figure \ref{fig:kinChain} (left).
The navigation frame $N$ is the global reference frame and its $z$-axis is aligned with gravity.
Each IMU $i \in \mathcal{I}, |\mathcal{I}| = N_I$ ($\mathcal{I}$ is the set of all IMU indices), has its own coordinate system, $I_i$, which is related to the navigation frame via a position (vector) $I^N_i \in \mathbb{R}^3$ and a rotation (matrix) $R^{NI_i} \in \mathcal{SO}(3)$. 
One IMU, typically the IMU at the chain root, is marked as $I_e$. 
Its role will be further explained in Section \ref{sec:problemStatement} below.
In general, we assume a kinematic chain with $|\mathcal{I}| = N_I \geq 2$ IMUs and $N_I-1$ joints, where the IMUs are mounted one-to-one on rigid segments connected by the joints. 
Throughout the paper we apply rotation matrices $\rot^{\ilocal\iimu_i}$, unit quaternions $\quat^{\ilocal\iimu_i}$ and Modified Rodriguez Parameters (\MRPs) $\mrp^{\ilocal\iimu_i}$ interchangeably to represent orientations.
Conversions between them are indicated as e.g. $q(R), \chi(q)$, see \cite{Markley2014}.
The position of a joint connecting two neighboring segments with rigidly attached IMUs $I_i$ and $I_j$ is denoted in each IMU coordinate frame as $J^{I_i}_{(i,j)}$ and $J^{I_j}_{(i,j)}, (i,j) \in \mathcal{J}$ ($\mathcal{J}$ is the set of all joint index pairs).
To describe the kinematic model, the global orientations of all IMUs in terms of \MRPs and the joint positions relative to the IMU coordinate frames (we call this IMU-centered joint positions) are used, as specified in Section \ref{sec:problemStatement}. 

\subsection{Estimation problem}
\label{sec:problemStatement}
We are interested in recursively estimating the time-dependent IMU angular kinematics and IMU-centered joint positions of a kinematic chain as outlined in the previous section. Note, that in contrast to \cite{Taetz2024},  we do not estimate the global chain position relative to the navigation frame and therefore we do not need position drift compensation. 
For each time instance $t$, we obtain the measurement vector $Y_t$ with synchronized measurements:
\begin{equation}
Y_t=\left(\begin{array}{c} \left\{\begin{array}{c} y^{I_i}_{a,t} \\ y^{I_i}_{\omega,t} \end{array}\right\}_{i\in \mathcal{I}} \\ y^{I_e}_{R,t} \end{array}\right),
\label{eqn:measuresSubnetwork}
\end{equation}
where $y^{I_i}_{a,t}, y^{I_i}_{\omega,t} \in \mathbb{R}^3$ denote the accelerometer and gyroscope measurement of IMU $i$. 
We assume all IMU measurements to be bias-free or at least bias compensated. 
For IMU $I_e$, we assume an additional measurement $y^{I_e}_{R,t}$ of its orientation relative to the navigation frame. 
The latter can be obtained from an external orientation estimator, such as \cite{Laidig_Seel_2022VQF}, or from another device.
The state $X_t$ at each time instance $t$ comprises:
\begin{equation}
    X_t = \left(
    \begin{array}{c}
    \left\{
    \begin{array}{c}   
    \chi^{NI_i} \\
    \omega^{NI_i}_{I_i}
    \end{array}
    \right\}_{i \in \mathcal{I}} \\
    \left\{
    \begin{array}{c}
    J^{I_i}_{(i,j)} \\
    J^{I_j}_{(i,j)}
    \end{array}
    \right\}_{(i,j) \in \mathcal{J}} 
    \end{array}
    \right)_t,
    \label{eqn:stateOrig}
\end{equation}
where $\chi^{NI_i}, \omega^{NI_i}_{I_i}, i\in \mathcal{I}$ denote the global IMU orientations and angular velocities, and $J^{I_i}_{(i,j)}, J^{I_j}_{(i,j)}, (i,j) \in \mathcal{J}$ are the IMU centered joint positions.
Note, adding the IMU angular velocities to the state, in contrast to \cite{weygersDriftFreeInertialSensorBased2020}, while computing the angular accelerations (which are also required for the estimation of the IMU-centered joint positions) based on these filtered quantities using finite differences, as specified in \eqref{eq:backward_diff} below, was found to be a good compromise between computational efficiency, drift reduction and accuracy. 

We can subdivide \eqref{eqn:stateOrig} into two parts, one having process noise in the time update and the other not, as:
\begin{equation}
 X_t = 
    \left\{ 
    \begin{array}{c}
        X^1_t \\
        X^2_t
    \end{array}
    \right\},
\end{equation}
where $X^1_t = \{ \omega^{NI_i}_{I_i}\}_{i \in \mathcal{I}}$ 
and $X^2_t$ covers the remaining components of $X_t$. 
With this splitting the state and the measurements are coupled via the discrete state space model:
\begin{align}
    X^1_t &= F^1_t(X_{t-1}) + w_t, \label{eqn:generalModel_top}\\
    X^2_t &= F^2_t(X_{t-1}), \label{eqn:motionModelConstraint}\\
    Y_t & = H_t(X_t) + v_t, \label{eqn:generalModel_bottom}
\end{align}
where $w_t \sim N(0, Q_t)$ and $v_t \sim N(0, \Sigma_t)$ denote independent zero mean Gaussian process and measurement noises.
Note that there is process noise only modeled for the states in $X^1_t$.
To incorporate the state transitions and measurements, we make use of a modified optimization-based recursive Bayesian filter as proposed in \cite{Taetz2024}.
Therefore, we perform an extended Kalman filter time udpate to obtain a prediction for $X_t \sim N(X_t;\hat{X}_{t|t-1}, \hat{P}_{t|t-1})$. Then, we solve the following maximum a-posteriori (MAP) problem:
\begin{align}
    \{\hat{X}_{t|t}, \hat{P}_{t|t} \} & = \min_{X} \frac{1}{2}\|\hat{X}_{t|t-1} - X \|_{\hat{P}_{t|t-1}^{-1}}^2+\sum_{i \in \mathcal{I}} \|v_t\|^2_{\Sigma^{-1}_t} 
    \label{eqn:CWeightedLeastSquares},
\end{align}
where $X_t \sim N(X_t;\hat{X}_{t|t}, \hat{P}_{t|t})$ is the state distribution of timestep $t$.
In the following, the time and measurement update models are introduced:
\subsubsection{Time update models $F_1$ and $F_2$}
\label{sec:ProcessModels}
The models with additive noise, which are used for $F^1_t$ in \eqref{eqn:generalModel_top}, assume constant angular velocity from time $t-1$ to $t$, with sampling time $\Delta t$. These are per IMU $i \in \mathcal{I}$:
\begin{align}
    \omega_{I_i,t}^{NI_i} &= \omega_{I_i,t-1}^{NI_i} + w^\omega_{t}, \label{eqn:stateSingle_end}
\end{align}
where $w^\omega_{t} \sim N(0, Q^\omega_{t})$.
The models without additive noise, which are used in $F^2_t$, are:
\begin{align}
    \chi^{NI_i}_{t} &= \chi\left( q^{NI_i}_{t-1} \odot \exp\left(\frac{\Delta t}{2} \omega_{I_i,t-1}^{NI_i} \right) \right),
\end{align}
for the global orientation of each IMU $i \in \mathcal{I}$, where $\odot$ indicates a quaternion multiplication, and
\begin{align}
    J^{I_i}_{(i,j),t} &= J^{I_j}_{(i,j),t-1},
    \label{eqn:jointPosIMU} 
\end{align}
for the IMU-centered joint positions of each joint index set $(i,j) \in \mathcal{J}$.

\subsubsection{Measurement update models $H_t$}
\label{sec:MeasModel}
The measured angular velocities are used to update the respective state variables per IMU $i \in \mathcal{I}$ with:
\begin{equation}
    y^{I_i}_{\omega,t} = \omega^{NI_i}_{I_i,t}+v^\omega_t,
\end{equation}
where $v^\omega_{t} \sim N(0, \Sigma^\omega_{t})$.
Moreover, as in \cite{weygersDriftFreeInertialSensorBased2020}, we exploit the fact that $\forall (i,j) \in \mathcal{J}$, the accelerations at joint position $J^{N}_{i,j}$, as measured from the two adjacent IMUs $I_i, I_j$, should coincide. 
This can be expressed per joint $(i,j) \in \mathcal{J}$ as:
\begin{equation}
R^{N I_i} a^{I_i}_{c,t} = R^{N I_j} a^{I_j}_{c,t} + v^a_{t}, \label{eq:JointAccContraint}
\end{equation}
where $v^a_{t} \sim N(0, \Sigma^a_t)$ denotes independent normally distributed measurement noise and $a^{I_i}_{c,t}, a^{I_j}_{c,t}$ are the IMU-centered accelerations at joint center $J_{(i,j),t}$. 
The latter are approximated via the acceleration measurements $y_{a,t}^{I_i}, y_{a,t}^{I_j}$ and the local joint position vectors $J^{I_i}_{(i,j),t}, J^{I_j}_{(i,j),t}$. 
For each IMU $i$ it holds:
\begin{align}
a_{c,t}^{I_i} &= y^{I_i}_{a,t} - \mathcal{C}_t^{I_i} J_{(i,j),t}^{I_i}, \\
\mathcal{C}_t^{I_i} &= \left[\omega_{I_i,t}^{NI_i} \times \right]^2 + \left[\dot{\omega}_{I_i,t}^{NI_i} \times \right].
\end{align}
IMU $j$ is handled analogously.
Here, $a\times$ denotes a cross product matrix of vector $a$ and angular acceleration $\dot{\omega}_{I_i,t}^{NI_i}$ is approximated via the following backward difference: 
\begin{align}
    \dot{\omega}_{I_i,t}^{NI_i} \approx \frac{y^{I_i}_{\omega,t} - \hat{\omega}_{I_i,t-1|t-1}^{NI_i}}{\Delta t},
    \label{eq:backward_diff}
\end{align}
with the sampling time $\Delta t$. 
Note, approximating the angular acceleration from angular velocity measurements alone did not result in accurate joint position estimates, and approximating the backward difference only from state variables did lead to drifting orientations in some of our pre-experiments. 
The proposed approximation in (\ref{eq:backward_diff}) showed good convergence of the joint position estimates and drift-free orientation estimates, as presented in the result section below, with reduced computational complexity compared to adding angular acceleration to the state.

The aforementioned models do not provide information about the global orientations of the IMUs relative to reference frame $N$, so that up to now, only drift-free relative orientations between IMUs could be estimated.
As shown in \cite{Lorenz_Taetz_Bleser_Stricker_2022}, the use of \eqref{eq:JointAccContraint} allows propagation of global orientation information through a kinematic chain, so that only one global orientation measurement $y^{NI_e}_{R,t}$ is needed to align all other IMUs of a kinematic chain with the navigation frame.
In order to achieve this, for the one IMU marked as $I_e$, the following measurement model is used with the external orientation measurement $y^{NI_e}_{R,t}$:
\begin{equation}
2 \log \left( \left(q\left(\chi^{NI_e}_t\right)\right)^{-1} \odot q\left(y^{NI_e}_{R,t}\right) \right) = v^R_{t},
\end{equation}
where $v^R_{t} \sim N(0, \Sigma^R_{t})$ and $\log$ denotes the quaternion logarithm. 

\subsection{Experimental setups}
\label{sec:Experimental_Setup}
As shown in \cite{kokObservabilityRelativeMotion2022}, in a magnetometer-free approach, specific motion excitation must occur for relative orientations between segments to become locally observable. This is fulfilled for our two different experimental setups illustrated in \figref{fig:kinChain}, which are described in the following.

\subsubsection{Manipulator study}
\label{sec:SimStudy}
To provide a first proof-of-concept we used a well-studied \cite{Taetz2016,Miezal2016,Lorenz_Taetz_Bleser_Stricker_2022,Taetz2024} manipulator simulation as shown on the left side of \figref{fig:kinChain}.
For each joint and each rotational degree of freedom a sinusoidal angle sequence using Denavit–Hartenberg parameters was defined as in \cite{Taetz2024}.
From this, we generated the \imu data with a custom built tool using forward kinematics.
The obtained data was sampled with 100~Hz. Additional i.i.d. zero-mean Gaussian noises, as observed from real IMUs, were added (with a variance of $8.25\cdot10^{-5}$ for the angular velocity values and a variance of $0.0075$ for the acceleration values).
The global orientation measurement, as ground truth orientation, was provided to the IMU with the least motion excitation, $I_0$.
We performed 100 Monte Carlo simulations each with a length of 10 minutes, with different state initializations. 
The lengths of the IMU-centered joint position vectors
were about 26 cm. 

\subsubsection{Lower body study}
\label{sec:Gait_setup}
To evaluate our approach on a biomechanically relevant movement, we used the publicly available TUK 6 minute walking dataset \cite{Teufl_Miezal_Taetz_Lorenz_Bleser-Taetz_2023}, for which the participants were equipped with seven XSens Awinda IMUs placed in custom 3D printed rigid bodies each carrying four reflective markers.
The rigid bodies were placed on the pelvis, thighs, shanks and feet using elastic straps.
The setup is illustrated on the right side of \figref{fig:kinChain}.
The participants walked on a track of about five meters back and forth for about six minutes.
From this dataset, we selected the first nine participants (seven female, two male, 20-25 years). 
In order to have high-quality ground truth data and reduce soft tissue and other artifacts in this study, we evaluated our approach on IMU data re-simulated from the optical reference.
Analog to the concept presented in \cite{mundtPredictionLowerLimb2020}, we simulated bias-free \imu data sampled with a frequency of 100~Hz based on the IMU poses from the optical reference, using similar noises as in the manipulator study.
The global orientation measurement, i.e. the orientation from the optical reference, was provided to the IMU on the pelvis.
The lengths of the IMU-centered joint position vectors were about 9 cm for the foot IMUs (to ankles) and ranged between 19 and 23 cm for the other IMUs. 

\section{Experimental Results}
\label{sec:results}
\begin{table}
	\centering
	\caption{\label{tab::noiseSettings}  Covariance matrices used for the optimization-based Bayesian Filter ($U$ is the unit matrix).} 
	\begin{tabular}{c  c  c  c  c c}
		$  P_{0,\chi}  $ & $P_{0,\omega}$ & $P_{0,J}$ & $Q^\omega_t $  & $\Sigma^{\omega}_t$ & $\Sigma^a_t$  
        \\
		\hline
		\rule{0pt}{8pt}
		$ 10^{-6} \, U$ & $ 10^{-1} U$ &  $10^{-4} U$  & $ 10^{-8} U$ & $10^{-3} U$ & $ 10^{-1} U$
	\end{tabular}
\end{table}
In order to assess our approach, we provide errors of the absolute orientation estimates of each IMU relative to the navigation frame, errors of the estimated relative IMU orientations (joint orientations) for each pair of IMUs connected by a joint and errors of the IMU-centered joint position estimates.
To assess the error in orientation we compute the error angle between a ground truth orientation $\bar{\rot}_{t}$ and the estimate $\hat{\rot}_{t|t}$ as
\begin{align}
	\alpha_\text{error,t} :=  \alpha\left( \bar{\rot}_{t} \; \hat{\rot}_{t|t}^{-1} \right),
\end{align}
where $\alpha\left( \cdot \right)$ denotes the extraction of the angle from an axis angle orientation representation.
For each IMU $i \in \mathcal{I}$, we consider the estimated absolute orientations $\hat{\rot}^{NI_i}_{t|t}$.
For each joint $(i,j) \in \mathcal{J}$ we consider the derived relative (joint) orientations $\hat{\rot}^{I_iI_j}_{t|t}$ between IMUs $I_i$ and $I_j$, with:
\begin{align}
	\hat{\rot}^{I_i,I_j}_{t|t} = \left(\hat{\rot}^{NI_i}_{t|t}  \right)^{-1} \hat{\rot}^{NI_j}_{t|t}.
\end{align}
An orientation drift would lead to an error increase over time. 
To test this, we split the time series of errors into batches of several minutes.
For each batch, we computed the mean average error (MAE) in orientation separately. 
Note that the error angles are always positive. 
Similar errors for subsequent batches speak for drift-free orientation estimates.
The error in the estimated joint positions is computed as follows.
Assume the ground truth joint position is given as $\bar{J}^{I_j}_{(i,j)}$ (analogously for $I_i$) and the estimate as $\hat{J}^{I_j}_{(i,j)}$, then the error in position $e_{\text{pos}, J^{I_j}_{(i,j)} }$ is defined as:
\begin{align}
	e_{\text{pos}, J^{I_j}_{(i,j)} } := ||\bar{J}^{I_j}_{(i,j)} \; - \; \hat{J}^{I_j}_{(i,j)}||_2 \;.
\end{align}

The covariance matrices used in the optimization-based Bayesian filter during our experimental evaluation are summarized in Table \ref{tab::noiseSettings}.
These were tuned manually, optimizing for drift-free absolute and joint orientations as well as joint positions. Moreover, we used randomly initialized joint positions as proposed in \cite{Taetz2024}.

\subsection{Manipulator study}
\begin{table}
\centering
\caption{Manipulator study: joint orientation and position errors \label{tab:JointManipulatorScenario}} 
\begin{tabular}{r | P{1.7cm} || l | p{1.3cm}   }
\hline
 \textbf{Joint} &  \textbf{Orientation error} [$\deg$] 
 & \textbf{IMU} 
 & \textbf{Position} \mbox{\textbf{error}}   [cm]   \\
\hline
\hline
\textbf{Joint(0,1)} &  0.25/0.012  \& & IMU0  &  1.53/0.012   \\
                &  0.26/0.016     & IMU1  &  0.25/0.008       \\
\hline
\textbf{Joint(1,2)} &  0.25/0.006  \& & IMU1  &  0.88/0.006   \\
                &  0.25/0.006     & IMU2  &  0.23/0.003      \\
\hline
\end{tabular}
\\
\vspace{0.1cm}
\footnotesize 
The entries are formatted as \textit{m(MAE)/std(MAE)}, where \textit{m(MAE)} is the median value over all MAEs for each trial and \textit{std(MAE)} is the standard deviation over all trials. In the gray colored column, the MAEs before the \textit{\&} indicate the first five minutes and after the last five minutes of the trials.
\vspace{-0.1cm}
\end{table}
\begin{table}
\centering
\caption{Manipulator study: global orientation errors \label{tab:AbsManipulatorScenario}} 
\begin{tabular}{r| p{1.45cm}  p{1.45cm}  p{1.45cm}  }
\hline
& \multicolumn{3}{c}{\textbf{Orientation error}  [$\deg$]}  \\
\textbf{IMU} &  1st 3.3min & 2nd 3.3min & 3rd 3.3min \\
\hline
IMU0 &  0.14/0.004  &  0.14/0.001  &  0.14/0.001  \\
IMU1 &  0.28/0.003  &  0.3/0.036  &  0.31/0.006  \\
IMU2 &  0.44/0.01  &  0.47/0.042  &  0.47/0.011  \\
\hline
\end{tabular}
\\
\vspace{0.1cm}
\footnotesize 
The entries are formatted in the same way as for Table~\ref{tab:JointManipulatorScenario}, while the trials have been divided into three batches to allow for more detailed analysis.
\vspace{-0.1cm}
\end{table}
The joint orientation errors are summarized in Table~\ref{tab:JointManipulatorScenario} (left).
All median joint orientation MAEs have very low values below 0.26 degrees with standard deviations below 0.016 degrees, indicating a drift-free estimation of the relative orientations.
The absolute orientation MAEs have median values lower than 0.47 degrees and are summarized in Table~\ref{tab:AbsManipulatorScenario}.
For the first 3.3 minutes, the errors are lower than for the second and third 3.3 minutes.
However, for the latter batches, the median MAEs do not significantly increase and the standard deviations even decrease from maximum 0.042 to 0.011 degrees.
Since we used the global orientation information for $I_0$, the errors are lowest for this IMU.
As in \cite{Lorenz_Taetz_Bleser_Stricker_2022}, IMUs further away from the chain root show increased absolute errors.
However, no orientation drift is present.

The joint position errors are summarized in Table~\ref{tab:JointManipulatorScenario} (right).
As expected, $J_{(0,1)}^{I_0}$, based on the IMU which perceives the least motion excitation, shows the highest median MAE of 1.53 cm, while $J_{(1,2)}^{I_2}$, based on the IMU with the highest motion excitation, shows the lowest median MAE of 0.23 cm. 
Note that all position estimates converged in less than 5 seconds and remained at a more or less static value during this experiment.
The average measured execution time per timestep (for three IMUs) was 0.51 ms  when running our single-threaded C++ implementation on an Intel i9-9820X (10 cores, Base Clock: 3.3 GHz).

\subsection{Lower body study}
\begin{table}
\centering
\caption{Lower body study: joint orientation and position errors \label{tab:JointLowerBodyScenario}} 
\begin{tabular}{r | P{1.7cm} || l | p{1.3cm}   }
\hline
 \textbf{Joint} & \textbf{Orientation error} [$\deg$] & \textbf{IMU} & \textbf{Position} \mbox{\textbf{error}} [cm]   \\
\hline
\hline
\textbf{L-Ankle} & 0.9/0.3 \& & L-Lo.Leg &  1.7/0.6    \\
                 & 0.8/0.3  & L-Foot &  1.1/0.2        \\
\hline
\textbf{L-Knee} & 0.9/0.3 \& & L-Up.Leg &  2.6/1.3  \\
                & 0.8/0.4 & L-Lo.Leg &  1.6/0.3     \\
\hline
\textbf{L-Hip} & 2.1/1.2 \& & Pelvis &  2.4/0.9   \\
                & 1.9/1.3 & L-Up.Leg &  2.0/0.3  \\
\hline
\textbf{R-Hip} &  2.6/1.0 \& & Pelvis &  2.7/1.0    \\
                & 2.5/0.6 & R-Up.Leg &  2.8/0.6     \\
\hline
\textbf{R-Knee} & 1.0/0.3  \& & R-Up.Leg &  1.9/1.2   \\
                    &  1.1/0.2& R-Lo.Leg &  1.4/0.3   \\
\hline
\textbf{R-Ankle} & 1.0/0.3 \& & R-Lo.Leg &  1.6/0.3   \\
                      & 0.9/0.3 & R-Foot &  1.1/0.2   \\
\hline
\end{tabular}
\\
\footnotesize 
The entries are formatted in the same way as for Table~\ref{tab:JointManipulatorScenario}, while the trials
have been divided into two batches of three minutes each.
\end{table}
\begin{table}
    \centering
    \caption{Lower body study: global orientation errors \label{tab:AbsLowerBodyScenario}} 
    \begin{tabular}{r| p{1.35cm}  p{1.35cm}   }
    \hline
    & \multicolumn{2}{c}{\textbf{Orientation error}  [$\deg$]}  \\
    \textbf{IMU} &  1st 3min &  2nd 3min \\
    \hline
    R-Up.Leg &  3.2/1.3  &  2.7/0.8  \\
    R-Lo.Leg &  3.3/1.4  &  2.5/0.9  \\
    R-Foot &  3.6/1.3  &  2.8/0.8  \\
    Pelvis &  1.8/0.9  &  1.4/0.5  \\
    L-Up.Leg &  2.3/1.7  &  2.2/1.4  \\
    L-Lo.Leg &  2.5/1.7  &  2.2/1.3  \\
    L-Foot &  2.4/1.6  &  2.0/1.3  \\
    \hline
    \end{tabular}
    \\
    \footnotesize 
    The entries are formatted in the same way as for Table~\ref{tab:JointManipulatorScenario}, while the trials
have been divided into two batches of three minutes each.
\end{table}
The results of the lower body study are summarized in Tables~ \ref{tab:JointLowerBodyScenario} (joint orientation and position errors) and~\ref{tab:AbsLowerBodyScenario} (global orientation errors).
Although the joint orientation errors in Table~ \ref{tab:JointLowerBodyScenario} (left) show a similar behavior as in the manipulator study, the median MAEs are up to two degrees higher.  
Since the pelvis IMU perceives the least motion excitation during walking, the joint orientation errors for the hips are, with a median MAE of 2.6 degrees for the right hip and 2.1 degrees for the left hip, the highest.
Knees and ankles show median MAEs ranging from 0.8 and 1.1 degrees.
The values for the first and second three minutes stay in the same ranges.
Except for the right knee, all median MAEs even decreased about 0.1 degrees, indicating a drift-free joint orientation estimation.
\begin{figure}
    \centering
    \includegraphics[width=0.6\columnwidth]{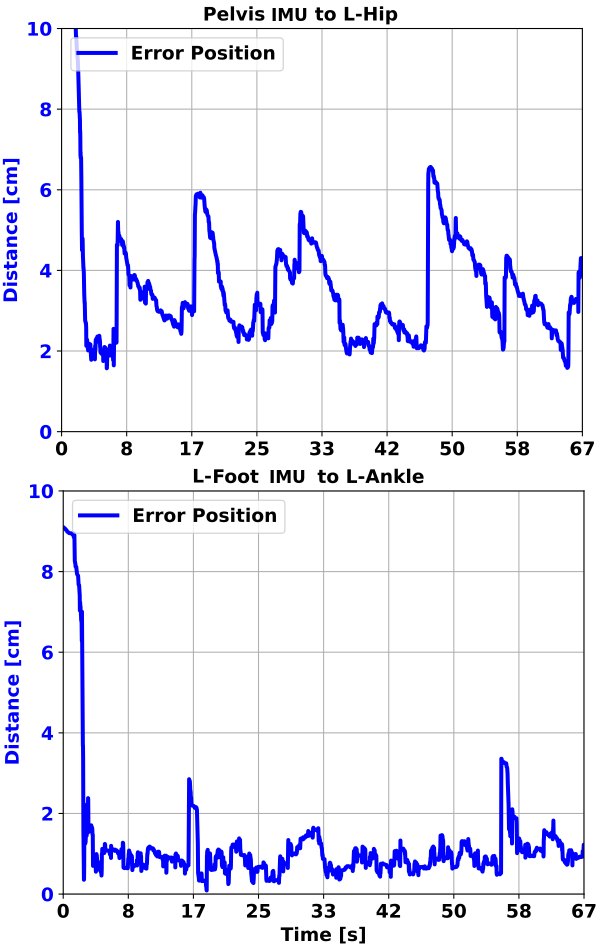}
    \caption{Selected joint position errors for one participant, first minute of the trial.
        \label{fig:ResultsPlot_JointPosition_LowerBody}}
    \vspace{-0.2cm}
 \end{figure}
The absolute orientation errors in Table~\ref{tab:AbsLowerBodyScenario} show the same behavior as in the manipulator study.
Since the global orientation information is used for the pelvis IMU, the pelvis also shows the lowest median MAEs for the absolute orientation in the first and second three minutes.
The IMUs of the right leg show median MAEs in a range from 3.2 to 3.6 degrees for the first three minutes and from 2.5 to 2.8 degrees for the second three minutes. 
These are slightly higher compared to the left leg with values ranging from 2.3 to 2.5 degrees for the first three minutes and 2.0 to 2.2 for the second three minutes.
A reason for this could be a higher motion excitation for the left leg IMUs during turning sequences.
Again, all IMUs show drift-free estimates for the absolute orientations, since the errors remain in constant ranges.

The joint position errors in Table~\ref{tab:JointLowerBodyScenario} (right) are higher compared to the manipulator study.
The lowest median MAE is 1.1 cm for the ankles relative to the corresponding foot IMUs.
The highest median MAEs are given for the hip joints ranging from 2.0 to 2.8 cm and for the position of the left knee relative to the corresponding left upper leg IMU with 2.6 cm, with overall standard deviations ranging from 0.2 to 1.3 cm.
To demonstrate the convergence behavior of the joint position estimation, \figref{fig:ResultsPlot_JointPosition_LowerBody} exemplifies the errors for the estimated positions of the left hip relative to the pelvis IMU (least motion excitation) and the left ankle relative to the left foot IMU (highest motion excitation) for one participant during the first minute of the trial. During the first 8 seconds, the position estimates converge and then vary depending on the motion excitation of the respective IMUs. As expected, the ankle related values show significantly less variation compared to the hip related values during the first minute of the trial.
The average measured execution time per timestep (for seven IMUs) was 5.6 ms, running on the same machine as for the previous study.

\section{Discussion}
\label{sec:discussion}
The experimental results demonstrate drift-free absolute and relative IMU orientation estimation of the proposed method.
In the manipulator study, all median MAEs of the orientations were below 0.47 degrees and below 2.8 degrees for the lower body study, after joint position convergence.
Specifically, errors remained stable and drift-free over extended durations, indicating successful implementation of the recursive Bayesian estimation with a minimal state representation.
The global orientation information was induced at IMU0 ($I_0$) for the manipulator study, and the pelvis IMU for the lower body study.
These IMUs showed the least global orientation errors compared to the other ones.
In addition, for the lower body study, hip joints associated with the pelvis IMU showed the highest relative orientation errors.
However, the latter can be explained by the fact that the pelvis IMU perceived less motion excitation according to the local observability criteria of~\cite{kokObservabilityRelativeMotion2022}.
This result aligns with previous studies such as~\cite{Lorenz_Taetz_Bleser_Stricker_2022}.

Furthermore, the experiments demonstrated acceptable online joint position estimates, with a median position MAE of at most 1.53 cm for the manipulator study and 2.8 cm for the lower body study.
The hip joint positions relative to the pelvis IMU showed comparably higher median position MAEs, which can again be explained by the lesser motion excitation of the pelvis IMU during walking.
In general, the errors related to the joint positions did not exhibit any systematic drift, affirming the algorithm's robustness and its potential applicability in real-world tracking scenarios where precise calibration is impractical, and a calibration-free method, such as the proposed one, is therefore preferable.
Moreover, the proposed method was shown to run in real-time (below IMU sampling rate) in the experimental evaluation (5.6 ms average measured processing time per timestep on 100 Hz data from seven IMUs).
Note, that the used C++ implementation is single-threaded and not optimized for speed: e.g. it uses numerical derivatives and no sparse matrices.

Finally, we would like to mention that the choice of covariance matrices and the initialization of the IMU orientations played an important role in obtaining accurate results. Hence, the sensitivity of the proposed algorithm with respect to these parameters should be further investigated. 

\section{Conclusions}
\label{sec:conclusions}
This work presented an online optimization-based Bayesian filtering approach for inertial tracking of kinematic chains.
Our method estimates not only the absolute orientations of IMUs attached to a kinematic chain, but also the joint positions relative to the IMU coordinate systems using a minimal state size. 
We evaluated our approach using both, simulated IMU data from a three-link kinematic chain (manipulator study) and IMU data re-simulated from a real lower body gait dataset (lower body study). 
The proposed algorithm demonstrated drift-free absolute and relative orientation estimates as well as joint position estimates with good accuracy across these different motion scenarios. 
The method shows high potential for real-world applications, providing a practical solution to calibration-free IMU-based motion tracking. 
Note, that based on the proposed approach, the kinematic chain pose can be illustrated by transforming the estimated IMU-centered joint positions into the navigation frame using the estimated global orientations and then connecting these.

\bibliographystyle{unsrt}  
\bibliography{references}

\begin{thebibliography}{10}

\bibitem{Cereatti2024}
Andrea Cereatti, Reed Gurchiek, Annegret Mündermann, Silvia Fantozzi, Fay Horak, Scott Delp, and Kamiar Aminian.
\newblock Isb recommendations on the definition, estimation, and reporting of joint kinematics in human motion analysis applications using wearable inertial measurement technology.
\newblock {\em Journal of Biomechanics}, 173:112225, 2024.

\bibitem{García-de-Villa2023}
Sara García-de Villa, David Casillas-Pérez, Ana Jiménez-Martín, and Juan~Jesús García-Domínguez.
\newblock Inertial sensors for human motion analysis: A comprehensive review.
\newblock {\em IEEE Transactions on Instrumentation and Measurement}, 72:1--39, 2023.

\bibitem{vitali2020}
Rachel~V. Vitali and Noel~C. Perkins.
\newblock Determining anatomical frames via inertial motion capture: {A} survey of methods.
\newblock {\em Journal of Biomechanics}, 106:109832, June 2020.

\bibitem{Seel2012}
Thomas Seel, Thomas Schauer, and Jörg Raisch.
\newblock Joint axis and position estimation from inertial measurement data by exploiting kinematic constraints.
\newblock In {\em 2012 IEEE International Conference on Control Applications}, pages 45--49, 2012.

\bibitem{Skoglund2015}
Martin Skoglund, Gustaf Hendeby, and Daniel Axehill.
\newblock Extended kalman filter modifications based on an optimization view point.
\newblock In {\em 18th International Conference of Information Fusion}, 07 2015.

\bibitem{olsson2017}
Fredrik Olsson and Kjartan Halvorsen.
\newblock Experimental evaluation of joint position estimation using inertial sensors.
\newblock In {\em 2017 20th {International} {Conference} on {Information} {Fusion} ({Fusion})}, pages 1--8, Xi'an, China, July 2017. IEEE.

\bibitem{weygersDriftFreeInertialSensorBased2020}
Ive Weygers, Manon Kok, Henri De~Vroey, Tommy Verbeerst, Mark Versteyhe, Hans Hallez, and Kurt Claeys.
\newblock Drift-{{Free Inertial Sensor-Based Joint Kinematics}} for {{Long-Term Arbitrary Movements}}.
\newblock {\em IEEE Sensors Journal}, 2020.

\bibitem{McGrath2021}
Timothy McGrath and Leia Stirling.
\newblock Body-worn imu human skeletal pose estimation using a factor graph-based optimization framework.
\newblock {\em Sensors}, 20(23), 2020.

\bibitem{Taetz2024}
B~Taetz, M~Lorenz, M~Miezal, D~Stricker, and G~Bleser-Taetz.
\newblock Jointtracker: Real-time inertial kinematic chain tracking with joint position estimation [version 2; peer review: 2 approved, 1 approved with reservations].
\newblock {\em Open Research Europe}, 4(33), 2025.

\bibitem{Lorenz_Taetz_Bleser_Stricker_2022}
Michael Lorenz, Bertram Taetz, Gabriele Bleser, and Didier Stricker.
\newblock Towards inertial human motion tracking with drift-free absolute orientations using only sparse sources of heading information.
\newblock In {\em Proceedings of the 25th International Conference on Information Fusion}, page~8, Linköping, Sweden, July 2022. IEEE.

\bibitem{Markley2014}
F.~Landis Markley and John~L. Crassidis.
\newblock {\em Fundamentals of Spacecraft Attitude Determination and Control}.
\newblock Springer, 2014.

\bibitem{Laidig_Seel_2022VQF}
Daniel Laidig and Thomas Seel.
\newblock Vqf: Highly accurate imu orientation estimation with bias estimation and magnetic disturbance rejection.
\newblock {\em Information Fusion}, 91:187--204, 2023.

\bibitem{kokObservabilityRelativeMotion2022}
Manon Kok, Karsten Eckhoff, Ive Weygers, and Thomas Seel.
\newblock Observability of the relative motion from inertial data in kinematic chains.
\newblock {\em Control Engineering Practice}, 125:105206, August 2022.

\bibitem{Taetz2016}
Bertram Taetz, Gabriele Bleser, and Markus Miezal.
\newblock Towards self-calibrating inertial body motion capture.
\newblock In {\em 19th International Conference on Information Fusion, {FUSION} 2016, Heidelberg, Germany, July 5-8, 2016}, pages 1751--1759. {IEEE}, 2016.

\bibitem{Miezal2016}
Markus Miezal, Bertram Taetz, and Gabriele Bleser.
\newblock On inertial body tracking in the presence of model calibration errors.
\newblock {\em Sensors}, 16:1132, 07 2016.

\bibitem{Teufl_Miezal_Taetz_Lorenz_Bleser-Taetz_2023}
Wolfgang Teufl, Markus Miezal, Bertram Taetz, Michael Lorenz, and Gabriele Bleser-Taetz.
\newblock Tuk-6-minute-walking dataset, December 2023.

\bibitem{mundtPredictionLowerLimb2020}
Marion Mundt, Wolf Thomsen, Tom Witter, Arnd Koeppe, Sina David, Franz Bamer, Wolfgang Potthast, and Bernd Markert.
\newblock Prediction of lower limb joint angles and moments during gait using artificial neural networks.
\newblock {\em Medical \& Biological Engineering \& Computing}, 58(1):211--225, January 2020.

\end{thebibliography}

\end{document}